\documentclass[letterpaper, 10 pt, conference]{ieeeconf}  

\IEEEoverridecommandlockouts                              

\usepackage{color}
\usepackage{colortbl}
\usepackage{xcolor}
\usepackage{graphicx}
\usepackage{hyperref}
\hypersetup{
    colorlinks=true,
    linkcolor=black,
    filecolor=magenta,
    urlcolor=blue,
    citecolor=black
}
\usepackage[acronym]{glossaries}

 \newcommand{\acronym}[1]{\gls{#1}\@}

\newacronym{fpfh}{FPFH}{Fast Point Feature Histogram}
\newacronym{icp}{ICP}{Iterative Closest Point}
\newacronym{iss}{ISS}{Intrinsic Shape Signature}
\newacronym{pca}{PCA}{Principal Component Analysis}
\newacronym{csg}{CSG}{Constrained Similarity Graph}
\newacronym{crf}{CRF}{Conditional Random Field}
\newacronym{board}{BOARD}{BOrder Aware Repeatable Directions}
\newacronym{tsdf}{TSDF}{Truncated Signed Distance Field}
\newacronym{ros}{ROS}{Robot Operating System}
\newacronym{gsm}{GSM}{Global Segmentation Map}
\newacronym{slam}{SLAM}{Simultaneous Localization and Mapping}
\newacronym{rmse}{RMSE}{Root-mean-square error}
\newacronym{ransac}{RANSAC}{Random Sample Consensus}
\newacronym{mav}{MAV}{Micro Air Vehicle}
\newacronym{uav}{UAV}{Unmanned Aerial Vehicle}
\newacronym{mavs}{MAVs}{Micro Air Vehicles}
\newacronym{uavs}{UAVs}{Unmanned Aerial Vehicles}
\newacronym{vio}{VIO}{Visual-Inertial Odometry}
\newacronym{viwls}{VIWLS}{Visual Inertial Weighted Least-Squares}
\newacronym{msf}{MSF}{Multi Sensor Fusion}
\newacronym{imu}{IMU}{Inertial Measurement Unit}
\newacronym{tnr}{T\&R}{Teach\&Repeat}
\newacronym{mpc}{MPC}{Model Predictive Controller}
\newcommand\blfootnote[1]{%
  \begingroup
  \renewcommand\thefootnote{}\footnote{#1}%
  \addtocounter{footnote}{-1}%
  \endgroup
}

\usepackage{xspace}
\makeatletter
\DeclareRobustCommand\onedot{\futurelet\@let@token\@onedot}
\def\@onedot{\ifx\@let@token.\else.\null\fi\xspace}

\let\NAT@parse\undefined
\makeatother

\usepackage[numbers,sort&compress]{natbib}

\definecolor{todo-red}{RGB}{200,12,12}
\definecolor{green4}{RGB}{0,128,0}

\title{\LARGE \bf
Visual-Inertial Teach and Repeat for Aerial Inspection
}

\author{Marius Fehr$^{1}$, Thomas Schneider$^{1}$, Marcin Dymczyk$^{1}$, J\"urgen Sturm$^{2}$ and Roland Siegwart$^{1}$
}

\begin{document}

\ieeefootline{Workshop on Aerial Robotic Inspection and Maintenance\\
International Conference on Robotics and Automation 2018, Brisbane, QLD, Australia}

\maketitle
\thispagestyle{empty} 

\footnotetext[1]{Autonomous Systems Lab, ETH Zurich, Switzerland.\\ {\tt\footnotesize \{marius.fehr, thomas.schneider, marcin.dymczyk\}@mavt.ethz.ch, rsiegwart@ethz.ch}}
\footnotetext[2]{Google Inc.: {\tt\footnotesize jsturm@google.com}}
\blfootnote{This reasearch was funded by Google Tango.}%


\begin{abstract}
Industrial facilities often require periodic visual inspections of key installations.
Examining these points of interest is time consuming, potentially hazardous or require special equipment to reach.
\acronym{mavs} are ideal platforms to automate this expensive and tedious task.
In this work we present a novel system that enables a human operator to teach a visual inspection task to an autonomous aerial vehicle by simply demonstrating the task using a handheld device.
To enable robust operation in confined, GPS-denied environments, the system employs the Google Tango visual-inertial mapping framework~\cite{google:tango} as the only source of pose estimates.
In a first step the operator records the desired inspection path and defines the inspection points.
The mapping framework then computes a feature-based localization map, which is shared with the robot.
After take-off, the robot estimates its pose based on this map and plans a smooth trajectory through the waypoints defined by the operator.
Furthermore, the system is able to track the poses of other robots or the operator, localized in the same map, and follow them in real-time while keeping a safe distance.
\end{abstract}

\section{INTRODUCTION}


Industrial facilities such as refineries, power and heating plants are required to operate at peak efficiency over extended periods of time.
Interruptions or failures due to wear and tear on the components of such installations are to be avoided at all cost, hence, regular inspections and revisions of key components are imperative.
However, inspecting these points of interest is time consuming and therefore expensive, especially if the installations need to be shut down due to potential or actual hazards to the technicians.
Other components might just simply be hard to reach and require special equipment.
Automating these inspection tasks using robots, such as~\acronym{mavs}, is a very cost-efficient solution, which can decrease down-times and most importantly keep the human work force out of harms way.

\begin{figure}[t]
\centering
\includegraphics[width=\columnwidth]{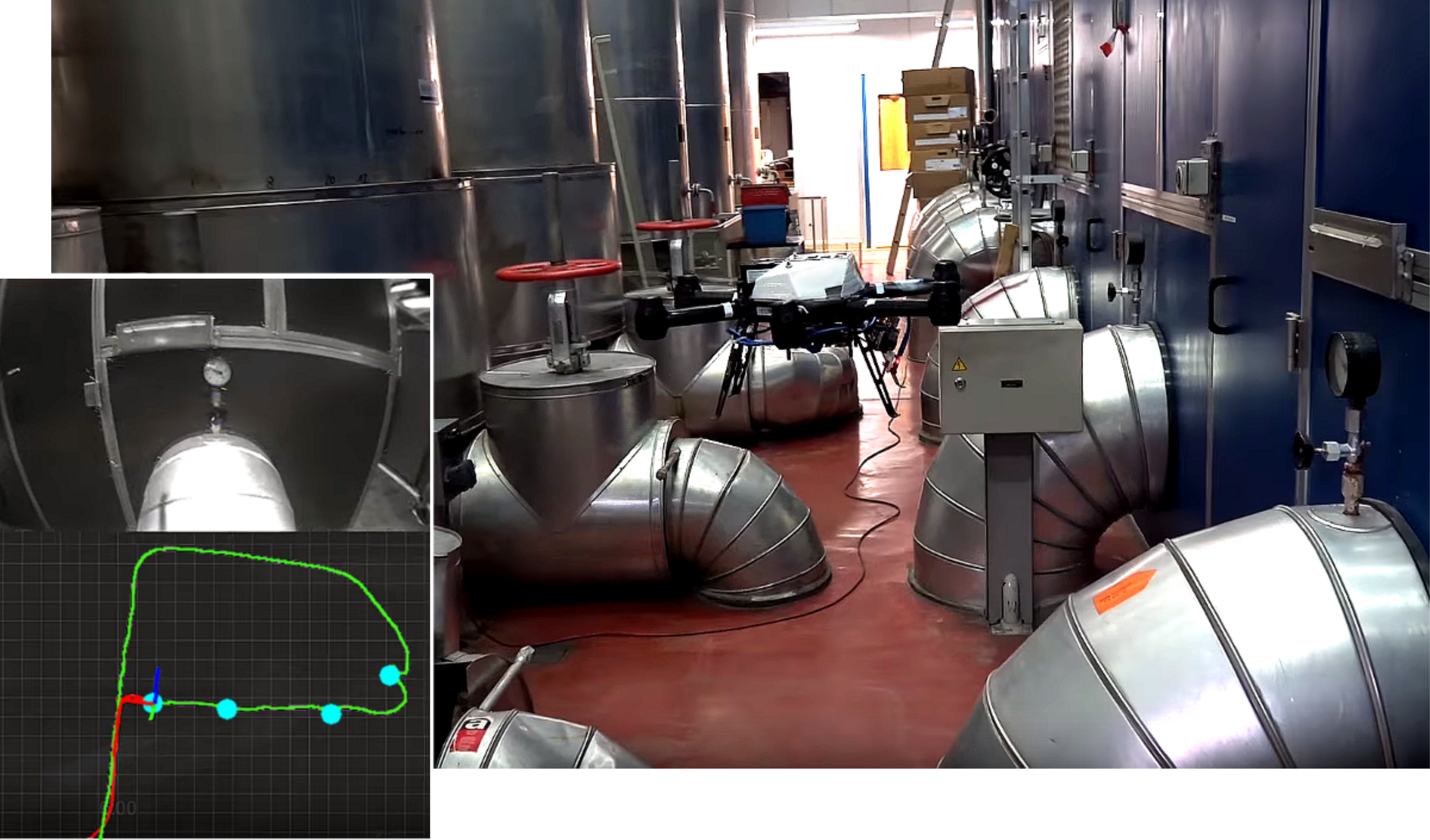}
\caption{The \acronym{mav} is autonomously performing an inspection task that has been taught by an operator holding a Google Tango tablet in a GPS-denied environment. See this video for the full demonstration of the system:~\url{https://goo.gl/V3TM6N}}
\label{fig:teaser}
\end{figure}

However, autonomous \acronym{mavs} require exact pose estimates to safely navigate in these potentially confined and GPS-denied environments.
That's why much of the industrial inspection research focuses either on the inspection task itself by providing change/damage detection algorithms~\cite{hallermann2014visual, morgenthal2014quality} or additional safety/navigation mechanisms to support the human pilot~\cite{eschmann2012unmanned, al2017vbii, sa2015outdoor}.
With the advent of robust \acronym{vio} pipelines and mapping framework, such as maplab~\cite{schneider2018maplab}, ORB-SLAM2~\cite{MurArtal2017}, OKVIS~\cite{leutenegger2015keyframe}, ROVIO~\cite{bloesch2015robust} and the visual-inertial navigation system presented in~\cite{lin2018autonomous} autonomous robots are able to perform more challenging tasks.
Without the need for external infrastructure, like cameras, markers or beacons, these systems allow the robots to follow waypoints provided by the operator in visually challenging environments~\cite{nikolic2013uav} or fully reconstruct the environment for the purpose of collision avoidance~\cite{omari2014visual}.

In this work we present a complete \acronym{tnr} system based on the Google Tango mapping framework~\cite{google:tango} that allows a non-expert human operator to simply demonstrate an industrial inspection task to the robot by simply walking and pointing using a hand-held tablet.
This allows the automation of tedious and potentially hazardous inspection routines in no more time than it takes to manually perform a single such task.
In case the inspection point is out of reach for the human operator, the inspection task can also be taught by manually piloting the \acronym{mav}.
The on-board \acronym{vio} allows the operator to navigate precisely and safely in position control mode, considerably lowering the need for extensive pilot training.

The contributions of this work are as follows:
\begin{itemize}
 \item We present a novel, robust visual-inertial-based \acronym{tnr} system for industrial inspection.
 \item We demonstrate the capabilities of the system in a challenging real-world industrial setting.
\end{itemize}

\section{SYSTEM}

\begin{figure}[t]
\centering
\includegraphics[width=0.9\columnwidth]{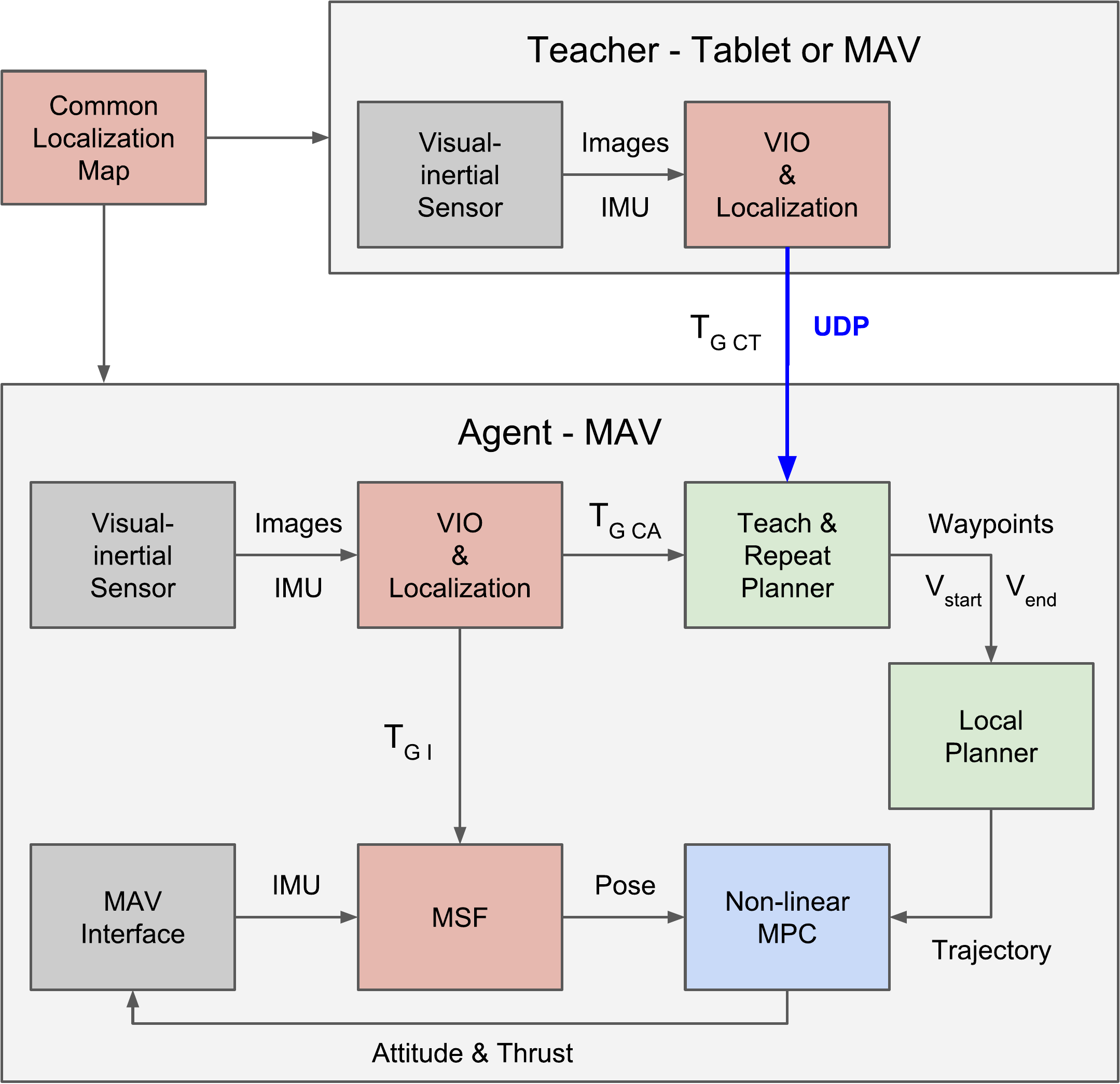}
\caption{\textbf{System overview:} Both teacher and agent (\acronym{mav}) are localizing from a local copy of a common feature-based localization map. The poses and inspection points of the teacher are sent to the agent live over a UDP connection and either followed or are recorded by the agent to execute the taught mission at any later point in time. The teacher poses and optional poses from other agents are furthermore used to prevent collisions. Components: \textit{red:} state estimation, \textit{green:} path planning, \textit{blue:} control, \textit{gray:} hardware}
\label{fig:overview}
\end{figure}

In this section we will introduce the architecture and components of the presented \acronym{tnr} system.
Fig.~\ref{fig:overview} provides an overview of the components and data flow of the proposed inspection system which is comprised of a teacher and an agent.
The teacher refers to a human operator either carrying a tablet or piloting an \acronym{mav}.
The agent on the other hand is a fully autonomous \acronym{mav}.
The system can be used in two different modes:
In \textit{live-mode} the \acronym{mav} will follow the operators poses in real-time, while keeping a safe distance in case it catches up.
In \textit{\acronym{tnr}-mode} the \acronym{mav} will record the operators poses and inspection points while on the ground and is able to then later execute the observed task as often as required.
In both modes the communication between teacher and agent was maintained over WiFi using UDP connection.\\
The remaining communication between the components as well as the components themselves are implemented based on the \acronym{ros}~\cite{ros2009}.

\textbf{State estimation: }
In order to be able to perfectly reenact the inspection task demonstrated by the teacher, both teacher and agent need to estimate their pose with respect to a common global frame.
This is achieved by means of robust \acronym{vio} and localization based on binary feature descriptors.
In order to create a localization map the initial map from odometry is post-processed using appearance-based loop closure and \acronym{viwls} optimization (See Fig.~\ref{fig:localization_map}).
Using this common localization map 6-DOF poses taught by the operator can be precisely mapped to the corresponding $[x,y,z,yaw]$ waypoints required by the \acronym{mav}.
On the \acronym{mav} the localized 6-DOF pose provided by the odometry is furthermore fused with the on-board \acronym{imu} for additional robustness using a \acronym{msf} based on~\cite{lynen13robust}.

\textbf{\acronym{tnr} planner: }
This global planner tracks the poses of the teacher and maintains a key-framed recording of its trajectory including the desired inspection points.
Based on the \acronym{mavs} position the planner will select a sequence of key-frames along the teachers trajectory.
There are two ways such a sequence can end, either in a normal key-frame, limited by a predefined maximum length or in an inspection point.
The planner will furthermore compute the desired velocity at the end of this sequence.
If the final waypoint is not the end of the trajectory and not equal to an inspection point this velocity is set to the maximum allowed velocity, otherwise the planner will set the velocity to zero.

\textbf{Safety: }
The sequence of waypoints and current agent position is continuously checked for collision against the safety spheres around both teacher and other agents.
If a sequence intersects with such a safety sphere, it is cut short and the terminal velocity set to zero, such that the agent comes to a full stop at a safe distance.
In case the agents own safety sphere intersects with another sphere, an emergency stop signal is sent directly to the controller, ordering an immediate stop of the \acronym{mav}.
It is important to note that aside from passive collision avoidance based on the known poses of the teacher and other agents, the system does not have any collision avoidance capabilities.
It therefore assumes a static environment with known dynamic obstacles.

\textbf{Local planner: }
The waypoints computed by the \acronym{tnr} planner are then used by the local planner to compute a smooth and dynamically feasible trajectory for the \acronym{mav} using polynomial trajectory optimization similar to~\cite{richter2016polynomial}.

\textbf{Control: }
Finally, the polynomial trajectory is sent to the non-linear \acronym{mpc}~\cite{kamelmpc2016}, which in turn computes the desired thrust and attitude signals for the \acronym{mav} interface.

\section{RESULTS}

\begin{figure}[t]
\centering
\includegraphics[width=\columnwidth]{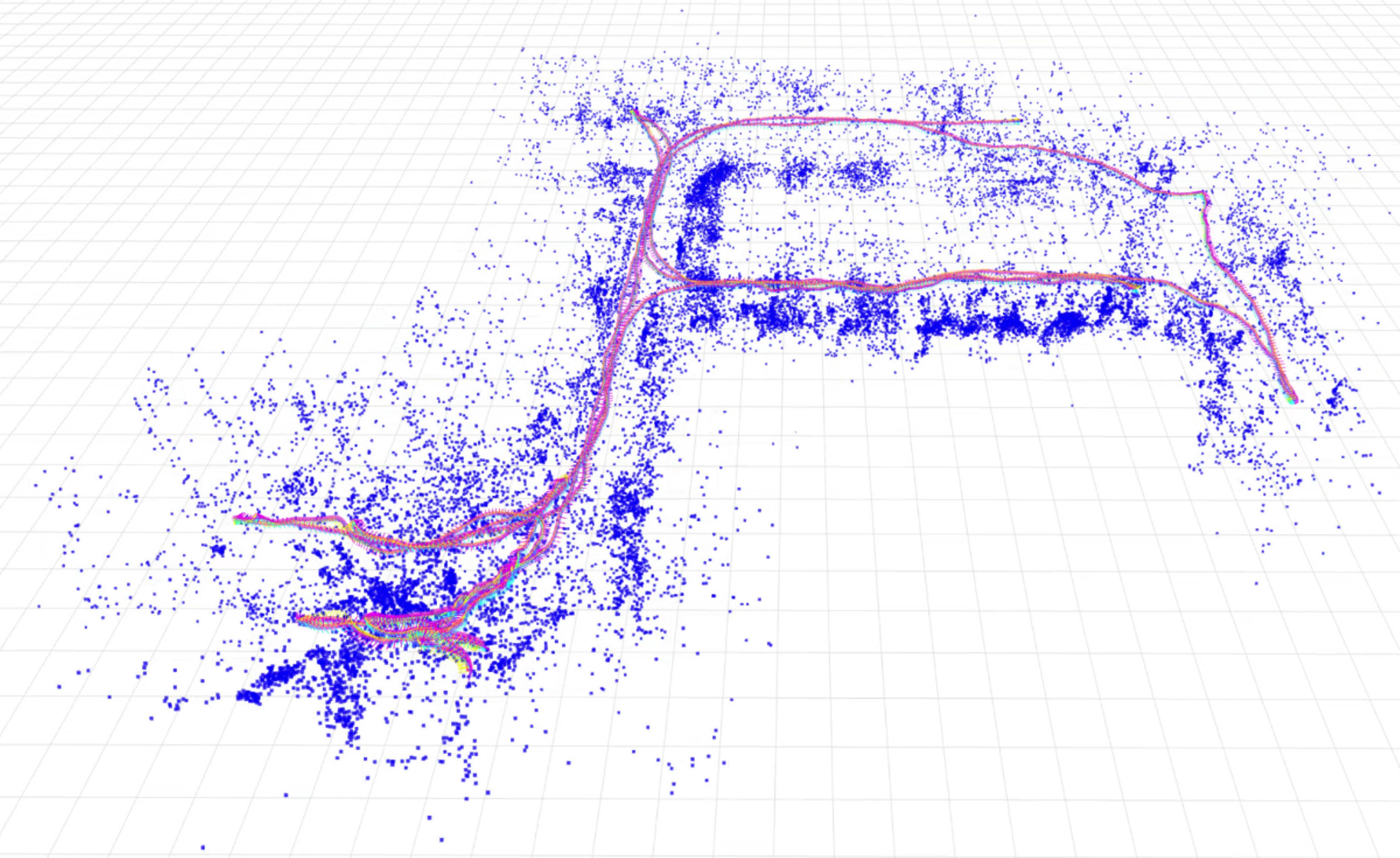}
\caption{Localization map of the industrial operating environment, created using the Google Tango framework~\cite{google:tango}.}
\label{fig:localization_map}
\end{figure}

\begin{figure}[h]
\centering
\includegraphics[width=\columnwidth]{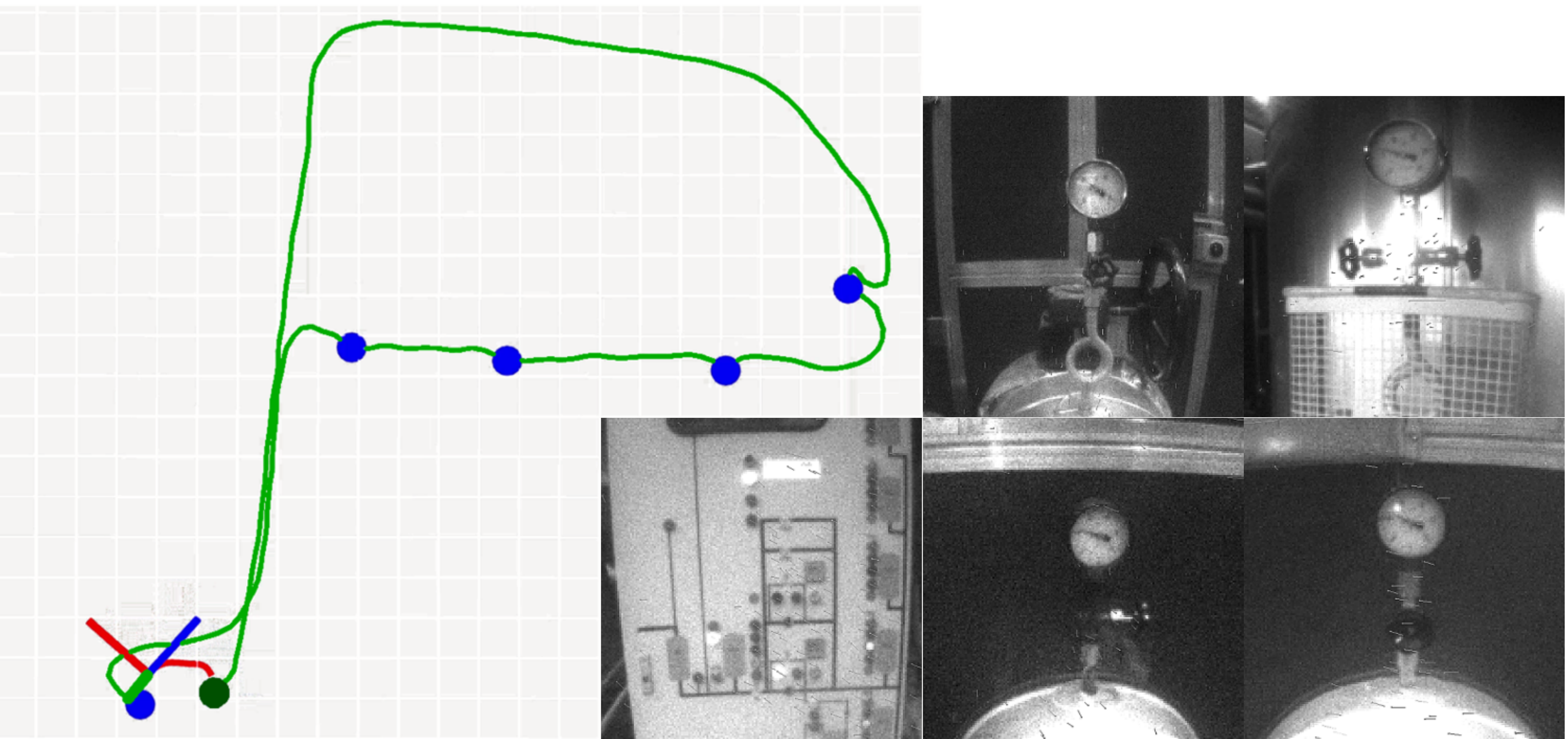}
\caption{The \acronym{mav} is executing the inspection plan (\textit{green:} trajectory, \textit{blue:} inspection points) and is on its way to the first inspection point. On the right are the resulting images of the 5 inspection points as seen by the \acronym{mav}.}
\label{fig:inspection_points}
\end{figure}

We demonstrate the capabilities of the proposed system in the following video:~\url{https://goo.gl/V3TM6N}.
The experiments were conducted on an AscTec Neo hexacopter, equipped with a visual-inertial stereo sensor and an Intel NUC, i7 on-board computer.
Prior to the experiments the operating environment was recorded using a Google Tango tablet.
The data was then downloaded to a external computer, loop-closed and bundle-adjusted and finally converted to a localization map using the Google Tango framework~\cite{google:tango}.
The resulting localization map can be seen in Fig.~\ref{fig:localization_map}.
At the beginning of the experiment the localization map was distributed among the teacher and the agent, i.e. the tablet and the \acronym{mav}.
All subsequent processing was executed exclusively on-board the tablet and the \acronym{mav}.
Localized to a common frame of reference, the inspection task is defined by the human operator using a hand-held Google Tango tablet.
Inspection points are automatically inserted if the operator remains motionless for 2 seconds.
The resulting inspection task can be seen in Fig.~\ref{fig:inspection_points}, where the \acronym{mav} has just started following the taught trajectory.
Furthermore, Fig.~\ref{fig:inspection_points} shows the inspected installations as seen from the \acronym{mav} after it safely and reliably completed the desired inspection task.

\section{CONCLUSION}

This work presents a novel system for autonomous industrial inspection using \acronym{mavs}, where the inspection task can be taught in an intuitive and safe manner by a human operator holding a tablet or piloting an \acronym{mav}.
In a GPS-denied environment, robust visual-inertial mapping techniques are used to localize teacher and agent in a common frame of reference, without the need for external infrastructure, such as markers or beacons.
To demonstrate the capabilities of the proposed system, we conducted a challenging real-world experiment in an industrial environment.
The \acronym{mav} followed the inspection plan safely and precisely and succeeded in observing all inspection points.
%
%





\section*{ACKNOWLEDGMENT}

The presented experiments received support from members of the Autonomous Systems Lab and Google Tango, most importantly: Michael Burri, Helen Oleynikova, Zachary Taylor, Fabian Bl\"ochliger, Mingyang Li, Ivan Dryanovski, Simon Lynen, and Konstantine Tsotsos.


\end{document}